\documentclass{article}
\usepackage{nips13submit_e,times}

\usepackage{natbib}
\usepackage{graphicx}
\usepackage{amsmath,amssymb,latexsym}

\usepackage{multirow}
\usepackage[applemac]{inputenc}

\newcommand{\bc}{\mbox{\boldmath $c$}}

\newcommand{\bh}{\mbox{\boldmath $h$}}

\newcommand{\bx}{\mbox{\boldmath $x$}}
\newcommand{\by}{\mbox{\boldmath $y$}}
\newcommand{\bz}{\mbox{\boldmath $z$}}

\newcommand{\bE}{\mbox{\boldmath $E$}}

\newcommand{\bW}{\mbox{\boldmath $W$}}

\newcommand{\balpha}{\mbox{\boldmath $\alpha$}}

\newcommand{\btheta}{\mbox{\boldmath $\theta$}}

\newcommand{\be}{\begin{eqnarray}}
\newcommand{\ee}{\end{eqnarray}}
\newcommand{\bee}{\begin{eqnarray*}}
\newcommand{\eee}{\end{eqnarray*}}

\newcommand{\matrixb}{\left[ \begin{array}}
\newcommand{\matrixe}{\end{array} \right]}

\usepackage{url}
\usepackage{amsmath}
\usepackage{amsthm}
\usepackage{amssymb}
\usepackage{graphicx}
\usepackage{xspace}
\usepackage{tabularx}
\usepackage{multicol}
\usepackage{multirow}
\usepackage{url}
\usepackage{wrapfig}
\usepackage[normalem]{ulem}
\usepackage{xcolor}

\newcommand{\RR}[0]{\mathbb{R}}

\newcommand{\NN}[0]{\ensuremath{\text{NN}}}
\newcommand{\TT}[0]{\ensuremath{\text{\btheta}}}

\newcommand{\ola}{\overleftarrow}
\newcommand{\ora}{\overrightarrow}

\title{Fine-Grained Attention Mechanism for Neural Machine Translation}
\author{Heeyoul Choi \\
Handong Global University\\
\texttt{heeyoul@gmail.com}
 \And Kyunghyun Cho \\
New York University\\
\texttt{ kyunghyun.cho@nyu.edu}\\
\ \And Yoshua Bengio\\
University of Montreal\\
CIFAR Senior Fellow\\
\texttt{yoshua.bengio@umontreal.ca}}

\nipsfinalcopy 

\begin{document}
\maketitle

\begin{abstract}
Neural machine translation (NMT) has been a new paradigm in machine translation, and the attention mechanism has become the dominant approach with the state-of-the-art records in many language pairs. While there are variants of the attention mechanism, all of them use only temporal attention where one scalar value is assigned to one context vector  corresponding to a source word. In this paper, we propose a fine-grained (or 2D) attention mechanism where each dimension of a context vector will receive a separate attention score. In experiments with the task of En-De and En-Fi translation, the fine-grained attention method improves the translation quality in terms of BLEU score.  In addition, our alignment analysis reveals how the fine-grained attention mechanism exploits the internal structure of context vectors. 
\end{abstract}

\section{Introduction}
\label{sec:introduction}

Neural machine translation (NMT), which is an end-to-end approach to machine
translation~\cite{kalchbrenner2013recurrent,Sutskever2014,Bahdanau2015},  has
widely become adopted in machine translation research, as evidenced by its
success in a recent WMT'16 translation
task~\cite{sennrich2016edinburgh,chung2016wmt}. The attention-based approach,
proposed by \cite{Bahdanau2015}, has become the dominant approach among
others, which has resulted in state-of-the-art translation qualities on, for
instance, En-Fr~\cite{Jean2015},
En-De~\cite{jean2015montreal,sennrich2016edinburgh}, En-Zh~\cite{Shen2016},
En-Ru~\cite{chung2016char} and
En-Cz~\cite{chung2016char,luong2016achieving}. These recent successes are
largely due to better handling a large target
vocabulary~\cite{Jean2015,Sennrich2015,chung2016char,luong2016achieving},
incorporating a target-side monolingual
corpus~\cite{sennrich2015improving,gulcehre2015using} and advancing the
attention
mechanism~\cite{luong2016effective,cohn2016incorporating,tu2016modeling}.

We notice that all the variants of the attention mechanism, including the
original one by \cite{Bahdanau2015}, are {\it temporal} in that it assigns a
scalar attention score for each context vector, which corresponds to a source
symbol. In other words, all the dimensions of a context vector are treated
equally. This is true not only for machine translation, but also for other tasks
on which the attention-based task was evaluated. For instance, the
attention-based neural caption generation by \cite{Xu2015show} assigns a scalar
attention score for each context vector, which corresponds to a spatial location
in an input image, treating all the dimensions of the context vector equally.
See \cite{cho2015describing} for more of such examples.

On the other hand, in~\cite{hchoi2017csl}, it was shown that word embedding vectors 
have more than one notions of similarities by analyzing the local chart of the manifold 
that word embedding vectors reside. Also, by contextualization of word embedding, 
each dimension of the word embedding vectors could play different role according to the context, 
which, in turn, led to better translation qualities in terms of the BLEU scores. 

Inspired by the contextualization of word embedding, 
in this paper, we propose to extend the attention mechanism so that each
dimension of a context vector will receive a separate attention score. 
This enables finer-grained attention, meaning that the attention mechanism may choose
to focus on one of many possible interpretations of a single word encoded in the
high-dimensional context vector~\cite{hchoi2017csl,van2012visualizing}. This is
done by letting the attention mechanism output as many scores as there are
dimensions in a context vectors, contrary to the existing variants of
attention mechanism which returns a single scalar per context vector.

We evaluate and compare the proposed fine-grained attention mechanism on the
tasks of En-De and En-Fi translation. The experiments reveal that the
fine-grained attention mechanism improves the translation quality up to +1.4 BLEU.
Our qualitative analysis found that the fine-grained attention mechanism indeed
exploits the internal structure of each context vector.

\section{Background: Attention-based Neural Machine Translation}

The attention-based neural machine translation (NMT) from \cite{Bahdanau2015} 
computes a
conditional distribution over translations given a source sentence $X=(w^x_1,
w^x_2, \ldots, w^x_T)$: 
\begin{align}
    \label{eq:cond_y_x}
    p(Y=(w^y_1, w^y_2, \ldots, w^y_{T'}) | X).
\end{align}
This is done by a neural network that consists of an encoder, a decoder and the
attention mechanism. 

The encoder is often implemented as a bidirectional recurrent neural network
(RNN) that reads the source sentence word-by-word. Before being read by the
encoder, each source word $w^x_t$ is projected onto a continuous vector space:
\begin{align}
    \label{eq:word_emb_x}
    \bx_t = \bE^x [\cdot, w^x_t],
\end{align}
where $\bE^x [\cdot, w^x_t]$ is $w^x_t$-th column vector of  $\bE_x \in \RR^{E
\times |V|}$, a source word embedding matrix, where $E$ and $|V|$ are the word
embedding dimension and the vocabulary size, respectively. 

The resulting sequence of word embedding vectors is then read by the
bidirectional encoder recurrent network which consists of forward and reverse
recurrent networks. The forward recurrent network reads the sequence in the
left-to-right order while the reverse network reads it right-to-left:
\begin{align*}
    \ora{\bh}_t = \ora{\phi}(\ora{\bh}_{t-1}, \bx_t),\\
    \ola{\bh}_t = \ola{\phi}(\ola{\bh}_{t+1}, \bx_t),
\end{align*}
where the initial hidden states $\ora{\bh}_0$ and $\ola{\bh}_{T+1}$ are
initialized as all-zero vectors or trained as parameters.  The hidden states
from the forward and reverse recurrent networks are concatenated at each time
step $t$ to form an annotation vector $\bh$:
\[
    \bh_t = \left[ \ora{\bh}_t; \ola{\bh}_t \right].
\]
This concatenation results in a context $C$ that is a tuple of annotation vectors:
\[
C = \left\{ \bh_1, \bh_2, \ldots, \bh_T \right\}.
\]
The recurrent activation functions $\ora{\phi}$ and $\ola{\phi}$ are in most
cases either long short-term memory units (LSTM, \cite{Hochreiter1997}) or gated
recurrent units (GRU, \cite{Cho2014}). 

The decoder consists of a recurrent network and the attention mechanism. The
recurrent network is a unidirectional language model to compute the conditional
distribution over the next target word given all the previous target words and
the source sentence:
\[
    p(w^y_{t'}|w^y_{<t'}, X).
\]
By multiplying this conditional probability for all the words in the target, we
recover the distribution over the full target translation in
Eq.~\eqref{eq:cond_y_x}.

The decoder recurrent network maintains an internal hidden state $\bz_{t'}$. At
each time step $t'$, it first uses the attention mechanism to select, or weight,
the annotation vectors in the context tuple $C$. The attention mechanism, which
is a feedforward neural network, takes as input both the previous decoder hidden
state, and one of the annotation vectors, and returns a relevant score $e_{t'
,t}$:
\begin{align}
    e_{t',t} = f_{\text{Att}}(\bz_{t'-1}, \bh_t),
    \label{eq:att0}
\end{align}
which is referred to as {\em score function} \cite{luong2016effective,chung2016char}. 
The function $f_{\text{Att}}$ can be implemented by fully connected neural networks with 
a single hidden layer where $tanh()$ can be applied as activation function. 
These relevance scores are normalized to be positive and sum to 1.
\begin{align}
    \label{eq:att_weight}
    \alpha_{t', t} = \frac{\exp(e_{t',t})}{\sum_{k=1}^T \exp(e_{t',k})}.
\end{align}
We use the normalized scores to compute the weighted sum of the annotation
vectors
\begin{align}
    \label{eq:wsum}
    \bc_{t'} = \sum_{t=1}^T \alpha_{t', t} \bh_t,
\end{align}
which will be used by the decoder recurrent network to update its own hidden
state by
\begin{align*}
    \bz_{t'} = \phi_z (\bz_{t'-1}, \by_{t'-1}, \bc_{t'}).
\end{align*}
Similarly to the encoder, $\phi_z$ is implemented as either an LSTM or GRU.
$\by_{t'-1}$ is a target-side word embedding vector obtained by
\begin{align*}
    \by_{t'-1} = \bE^y [\cdot, w^y_{t'-1}],
\end{align*}
similarly to Eq.~\eqref{eq:word_emb_x}.

The probability of each word $i$ in the target vocabulary $V'$ is computed by
\begin{align*}
    p(w^y_{t'}=i | w^y_{<t'}, X) = \phi \left( 
        \bW^y_i \bz_{t'} + c_i
    \right),
\end{align*}
where $\bW^y_i$ is the $i$-th row vector of $\bW^y \in  \RR^{ |V| \times
dim(\bz_{t'})} $ and $c_i$ is the bias.

The NMT model is usually trained to maximize the
log-probability of the correct translation given a source sentence using a large
training parallel corpus. This is done by stochastic gradient descent, where the
gradient of the log-likelihood is efficiently computed by the backpropagation
algorithm.

\subsection{Variants of Attention Mechanism}

Since the original attention mechanism was proposed as in Eq.~\eqref{eq:att0}
\cite{Bahdanau2015}, there have been several variants
\cite{luong2016effective}.

\cite{luong2016effective} presented a few variants of the attention 
mechanism on the sequence-to-sequence model \cite{Sutskever2014}.
Although their work cannot be directly compared to the attention model in \cite{Bahdanau2015}, 
they introduced a few variants for score function of attention model--content based 
and location based score functions. Their score functions still assign a single value 
for the context vector $\bh_t$ as in Eq.~\eqref{eq:att0}. 

Another variant is to add the target word embedding 
as input for the score function \cite{Jean2015,chung2016char} as follows: 
\begin{align}
    e_{t',t} = f_{\text{AttY}}(\bz_{t'-1}, \bh_t, \by_{t'-1}),
    \label{eq:att_y}
\end{align}
and the score is normalized as before, which leads to $\alpha_{t',t}$, and $f_{\text{AttY}}$ can be a
fully connected neural network as Eq. (\ref{eq:att0}) with different input size. 
This method provides the score  
function additional information from the previous word. In training, teacher forced 
true target words can be used, while in test the previously generated word is used. 
In this variant, still a single score value is given to the context vector $\bh_t$. 


\section{Fine-Grained Attention Mechanism}

All the existing variants of attention mechanism assign a single scalar score
for each context vector $\bh_t$.
We however notice that it is not necessary to assign a single score to the
context at a time, and that it may be beneficial to assign a score for each {\it
dimension} of the context vector, as each dimension represents a different
perspective into the captured internal structure. In \cite{hchoi2017csl}, 
it was shown that each dimension in word embedding could have different meaning 
and the {\em context} could enrich the meaning of each dimension in different ways. 
The insight in this paper is similar to \cite{hchoi2017csl}, except two points: 
(1) focusing on the encoded representation rather than word embedding, 
and (2) using 2 dimensional attention rather than the context of the given sentence. 


We therefore propose to extend the score function $f_{\text{Att}}$ in
Eq.~\eqref{eq:att0} to return a set of scores corresponding to the dimensions of
the context vector $\bh_t$. That is,
\begin{align}
    e^{d}_{t',t} = f^{d}_{\text{AttY2D}}(\bz_{t'-1}, \bh_t, \by_{t'-1}),
    \label{eq:att2_y}
\end{align}
where $e^d_{t',t}$ is the score assigned to the $d$-th dimension of the $t$-th
context vector $\bh_t$ at time $t'$. Here, $f_{\text{AttY2D}}$ is a fully connected neural network 
where the number of output node is $d$.  
These dimension-specific scores are further
normalized dimension-wise such that 
\begin{align}
    \label{eq:att2y_weight}
    \alpha^d_{t', t} = \frac{\exp(e^d_{t',t})}{\sum_{k=1}^T \exp(e^d_{t',k})}.
\end{align} 
The context vectors are then combined by
\begin{align}
    \label{eq:att2y_wsum}
    \bc_{t'} = \sum_{t=1}^T \balpha_{t', t} \odot \bh_t, 
\end{align}
where $\balpha_{t',t}$ is $\left[ \alpha^1_{t',t}, \ldots,
\alpha^{\dim(\bh_t)}_{t',t}\right]^\top$, and $\odot$ an element-wise
multiplication.

%

\begin{figure}[t]
        \centering{\includegraphics[width=3.5in]{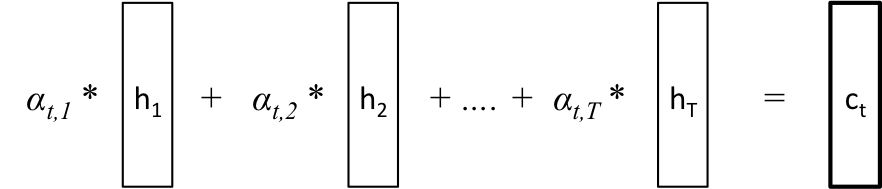}}
        
        \centering{(a)}
        \vspace{0.3in}

        \centering{\includegraphics[width=3.5in]{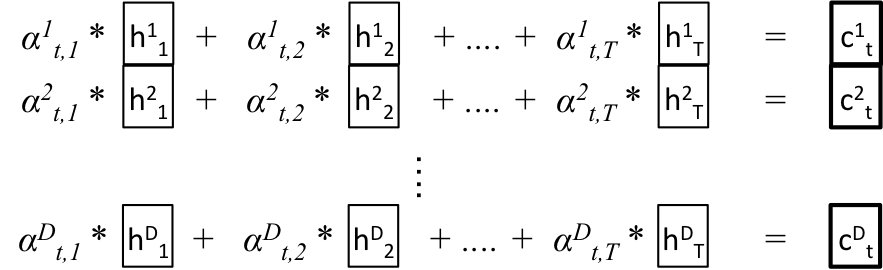}}

        \centering{(b)}

    \caption{(a) The conventional attention mechanism and (b) The proposed
    fine-grained attention mechanism. Note that $\sum_t \alpha_{t', t} = 1$ in the 
    conventional method, and $\sum_t \alpha^d_{t', t} = 1$ for all dimension $d$ in 
    the proposed method.}
\label{fig:att_models}
\end{figure}

We contrast the conventional attention mechanism against the proposed
fine-grained attention mechanism in Fig.~\ref{fig:att_models}.

\section{Experimental Settings}

\subsection{Tasks and Corpora}

We evaluate the proposed fine-grained attention mechanism on two translation
tasks; (1) En-De and (2) En-Fi. For each language pair, we use all the parallel
corpora available from WMT'15\footnote{
    \url{http://www.statmt.org/wmt15/}
}
for training, which results in 4.5M and 2M sentence pairs for En-De and En-Fi,
respectively. In the case of En-De, we preprocessed the parallel corpora
following \cite{Jean2015} and ended up with 100M words on the English side. For
En-Fi, we did not use any preprocessing routine other than simple tokenization.

Instead of space-separated tokens, we use 30k subwords extracted by byte pair
encoding (BPE), as suggested in \cite{Sennrich2015}. When computing the
translation quality using BLEU, we {\it un-BPE} the resulting translations, but
leave them tokenized.

\subsection{Decoding and Evaluation}

Once a model is trained, we use a simple forward beam search with width set to
12 to find a translation that approximately maximizes $\log p(Y|X)$ from
Eq.~\eqref{eq:cond_y_x}. The decoded translation is then un-BPE'd and evaluated
against a reference sentence by BLEU (in practice, BLEU is computed over a set
of sentences.) We use newstest2013 and newstest2015 as the validation and test
sets for En-De, and newsdev2015 and newstest2015 for En-Fi. 

\subsection{Models}

We use the attention-based neural translation model from \cite{Bahdanau2015} as
a {\bf baseline}, except for replacing the gated recurrent unit (GRU) with the
long short-term memory unit (LSTM). 
The vocabulary size is 30K for both source and target languages, 
the dimension of word embedding is 620 for both languages, 
the number of hidden nodes for both encoder and decoder is 1K,
and the dimension of hidden nodes for the alignment model is 2K.
 
Based on the above model configuration, we
test a variant of this baseline model, in which we feed the previously decoded
symbol $y_{t-1}$ directly to the attention score function $f_{\text{Att}}$ from
Eq.~\eqref{eq:att0} ({\bf AttY}). These models are compared against the model
with the proposed fine-grained model ({\bf AttY2D}).

We further test adding a recently proposed technique, which treats each dimension of word embedding differently based on the context. This looks similar to our fine-grained attention in a sense that each dimension of the representation is treated in different ways. We evaluate the
contextualization ({\bf Context}) proposed by \cite{hchoi2017csl}. The
contextualization enriches the word embedding vector by incorporating the
context information:
\begin{align*}
    \bc^x = \frac{1}{T} \sum_{t=1}^T \NN_{\TT}(\bx_t),
\end{align*}
where $\NN_{\TT}$ is a feedforward neural network parametrized by $\TT$. We
closely follow \cite{hchoi2017csl}.

All the models were trained using Adam~\cite{Kingma2014adam} until the BLEU
score on the validation set stopped improving. For computing the validation
score during training, we use greedy search instead of beam search in order to
minimize the computational overhead. That is 1 for the beam search. 
As in \cite{Bahdanau2015}, we trained our model with the sentences of length up to 50 words.

\begin{table}[t]
    \centering
\begin{tabular}{l||c|c|c|c|c|c}
& \multicolumn{2}{c|}{En-De} & \multicolumn{2}{c|}{En-Fi} \\
\hline
Beam Width 	&  1 & 12 			& 1 & 12 \\ 
\hline
\hline
\small Baseline		&   {\small 17.57 (17.62)}&  {\small 20.78 (19.72)} 	&  {\small 6.07 (7.18)}&  {\small 7.83 (8.35)}\\
\small +AttY		&  {\small 19.15 (18.82)}&  {\small 21.41 (20.60)}  	&  {\small 7.38 (8.02)}&  {\small 8.91 (9.20)}\\
\small +AttY2D         &  {\small 20.49 (19.42)}&  {\small 22.50 (20.83)}    	&  {\small 8.33 (8.75)}&  {\small 9.32 (9.41)}\\
\small +Context(C)  &  {\small 19.13 (18.81)}&  {\small 22.13 (21.01)} 		&  {\small 7.47 (7.93)}&  {\small 8.84 (9.18)}\\
\small +C+AttY    &  {\small 20.96 (20.06)}&  {\small 23.25 (21.35)}   &  {\small 8.67 (9.18)}&   {\small 10.01 (9.95)} \\
\small +C+AttY2D    &  {\small 22.37 (20.56)}&  {\small {\bf 23.74} (22.13)}   &  {\small 9.02 (9.63)}&  {\small {\bf 10.20} (10.90)} \\
\hline
\hline
\end{tabular}
\caption{BLEU scores on the test sets for En-De and En-Fi with two different
beam widths. The scores on the development sets are in the parentheses. 
The baseline is the vanilla NMT model from \cite{Bahdanau2015} with 
LSTM and BPE.}
\label{table:bleu}
\end{table}%

\section{Experiments}

\subsection{Quantitative Analysis}

We present the translation qualities of all the models on both En-De and En-Fi
in Table~\ref{table:bleu}. We observe up to +1.4 BLEU when the proposed
fine-grained attention mechanism is used instead of the conventional attention
mechanism ({\bf Baseline} vs {\bf Baseline}+{\bf AttY} vs {\bf Baseline}+{\bf
AttY2D}) on the both language pairs. These results clearly confirm the
importance of treating each dimension of the context vector separately. 

With the contextualization (+{\bf Context} or +{\bf C} in the table), we observe the same 
pattern of improvements by the proposed method. Although the contextualization 
alone improves BLEU by up to +1.8 compared to the baseline, 
the fine-grained attention boost up the BLEU score by additional +1.4. 

The improvements in accuracy require additional time as well as larger model size. 
The model size increases 3.5\% relatively from {\bf +AttY} to {\bf +AttY2D}, 
and 3.4\% from {\bf +C+AttY} to {\bf +C+AttY2D}. 
The translation times are summarized in 
Table. \ref{table:time}, which shows the proposed model needs extra time (from 4.5\% to 14\% relatively).

\begin{table}[h]
    \centering
\begin{tabular}{l||l|l|}
Models & {En-De} & {En-Fi} \\
\hline
\hline
\small Baseline+AttY		     & \small   2,546 &  \small 1,631 \\
\small Baseline+AttY2D         &  \small  2,902 (+14.0\%) &  \small  1,786 (+9.5\%) \\
\hline
\small Baseline+C+AttY         &  \small  2,758 &  \small  1,626 \\
\small Baseline+C+AttY2D    &  \small  2,894 (+4.5\%) &  \small  1,718 (+5.7\%) \\
\hline
\hline
\end{tabular}
\caption{Elapsed time (in seconds) for translation of test files. 
The test file `newstest2015' for En-De has 2,169 sentences and 
`newstest2015' for En-Fi has 1,370 sentences. 
The numbers in the parenthesis indicate the additional times for {\bf AttY2D}
compared to the corresponding {\bf AttY} models. }
\label{table:time}
\end{table}%


\subsection{Alignment Analysis}

Unlike the conventional attention mechanism, the proposed fine-grained one
returns a 3--D tensor $\alpha^d_{t',t}$ representing the relationship between the
triplet of a source symbol $x_{t}$, a target symbol $y_{t'}$ {\it and} a
dimension of the corresponding context vector $\bc_t^d$. This makes it
challenging to visualize the result of the fine-grained attention mechanism,
especially because the dimensionality of the context vector is often larger (in
our case, 2000.) 

Instead, we first visualize the alignment averaged over the dimensions of a
context vector:
\[
    A_{t,t'} = \frac{1}{\dim(\bc_t)} \sum_{d=1}^{\dim(\bc_t)} \alpha^d_{t',t}.
\]
This computes the {\it strength} of alignment between source and target symbols,
and should be comparable to the alignment matrix from the conventional attention
mechanism.

\begin{figure}[t]
        \centering{
        \includegraphics[height=1.85in]{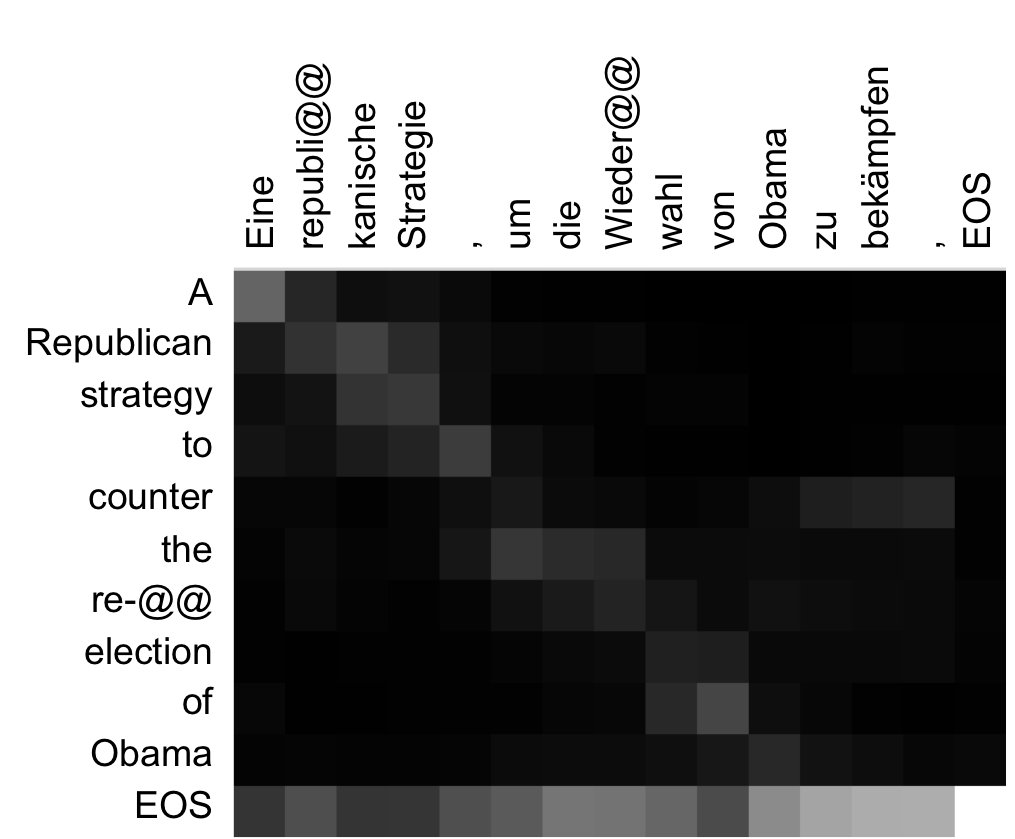}
        \includegraphics[height=1.85in]{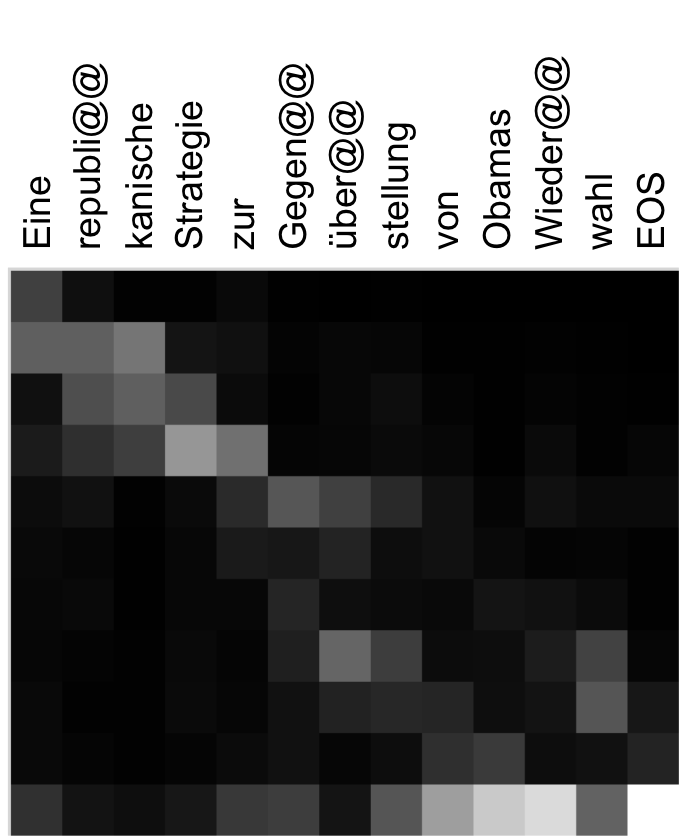}
        \includegraphics[height=1.85in]{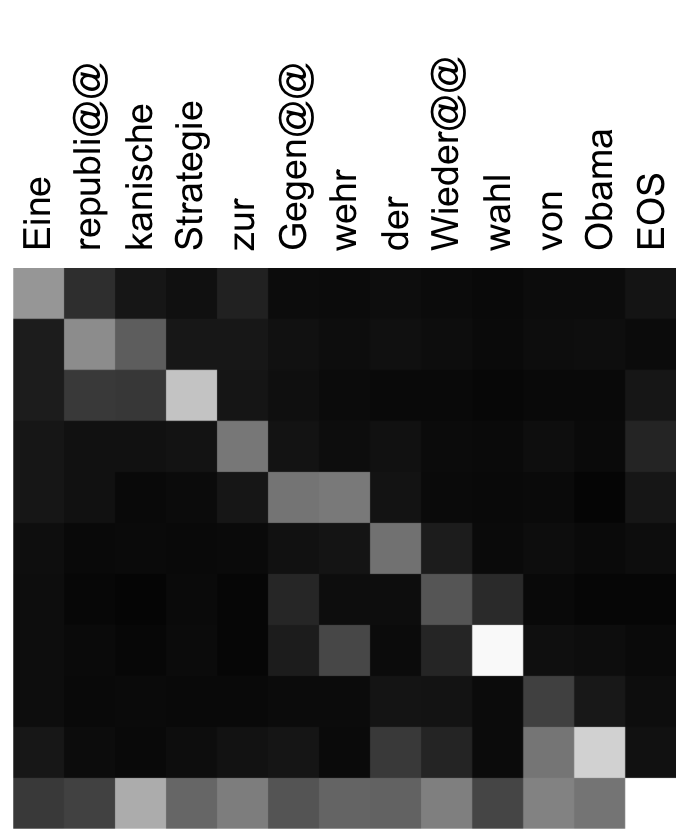}}

        \centering{\hspace{0.7in}(a) \hspace{1.4in} (b) \hspace{1.20in} (c)}
    \caption{Attention assignments with different attention models in 
    the En-De translation: 
    (a) the vanilla attention model (Att), 
    (b) with target words $\by_{t'-1}$ (AttY), and 
    (c) the proposed attention model (AttY2D).}
\label{fig:att}
\end{figure}

In Fig.~\ref{fig:att}, we visualize the alignment found by (left) the original
model from \cite{Bahdanau2015}, (middle) the modification in which the
previously decoded target symbol is fed directly to the conventional attention
mechanism ({\bf AttY}), and (right) the averaged alignment $A_{t,t'}$ from the
proposed fine-grained attention mechanism. There is a clear similarity among
these three alternatives, but we observe a more clear, focused alignment in the
case of the proposed fine-grained attention model. 

Second, we visualize the alignment averaged over the target:
\[
    A_{t,d} = \frac{1}{|Y|} \sum_{t'=1}^{|Y|} \alpha^d_{t',t}.
\]
This matrix is expected to reveal the dimensions of a context vector per source
symbol that are relevant for {\it translating} it without necessarily specifying
the aligned target symbol(s). 

\begin{figure}[h]
        \centering{\includegraphics[width=5.33in]{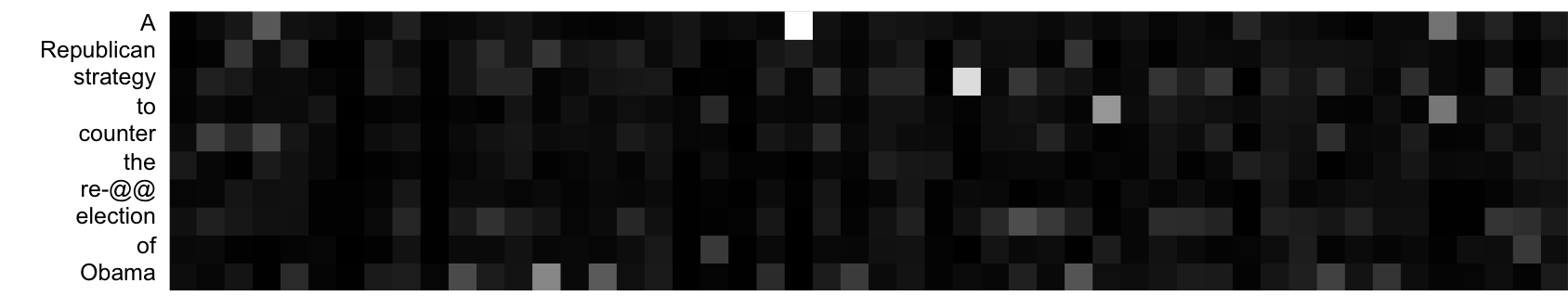}}
    \caption{Attention assignments with the fine-grained attention model. Due to 
    the limit of the space, only the first 50 dimensions are presented. The vertical 
    and the horizontal axes indicate the source sub-words and the 50 dimensions 
    of the context vector $\bh_t$, respectively.}
\label{fig:att_dims}
\end{figure}

In Fig.~\ref{fig:att_dims}, we can see very sparse representation where each source 
word receives different pattern of attentions on different dimensions.

\begin{figure}[h]
        \centering{
        \includegraphics[height=1.9in]{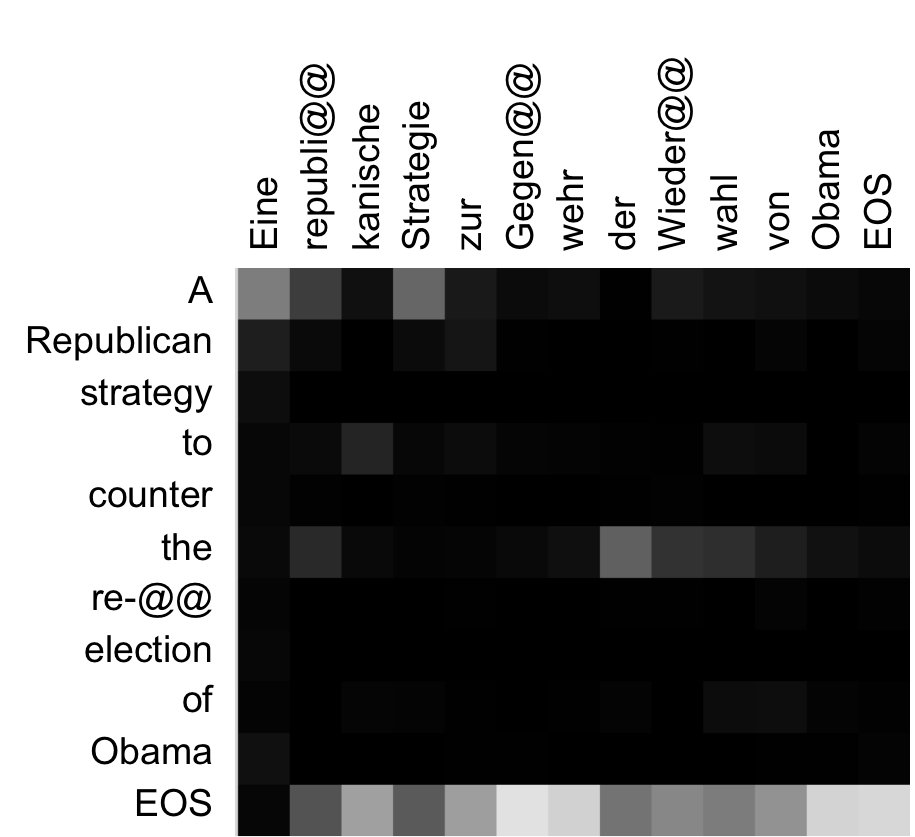}
        \includegraphics[height=1.9in]{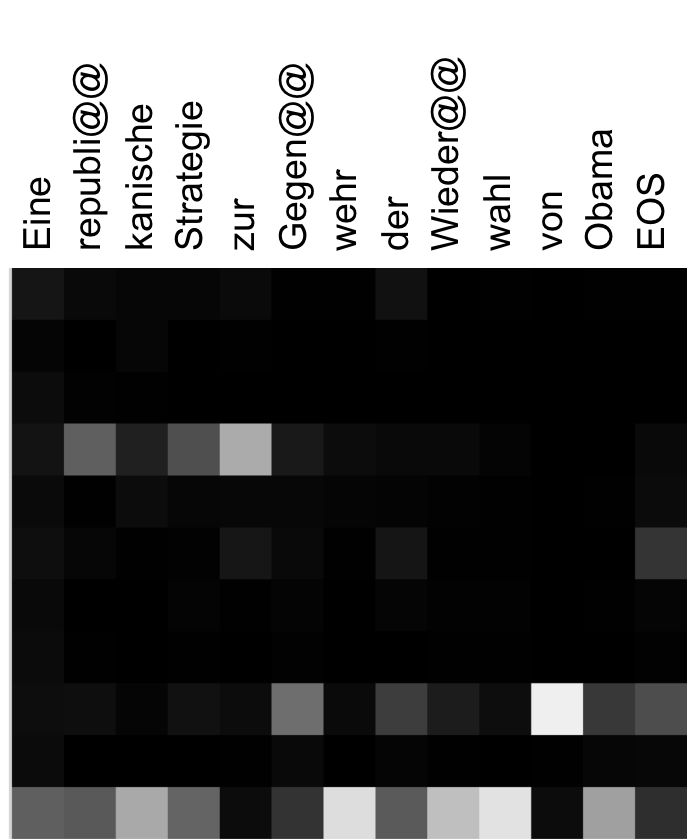}
        \includegraphics[height=1.9in]{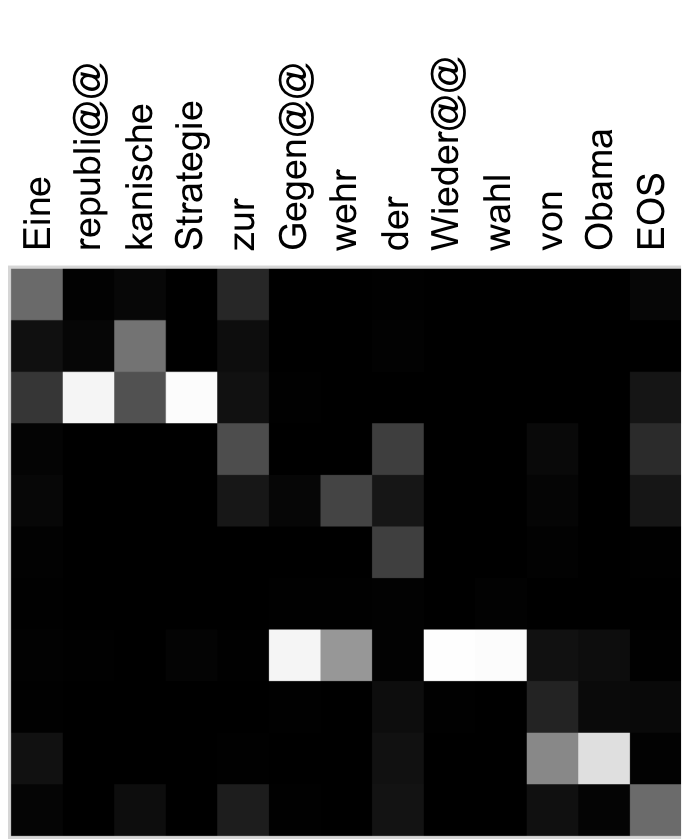}}

        \centering{\hspace{0.7in}(a) \hspace{1.40in} (b) \hspace{1.20in} (c)}

        \centering{
        \includegraphics[height=1.9in]{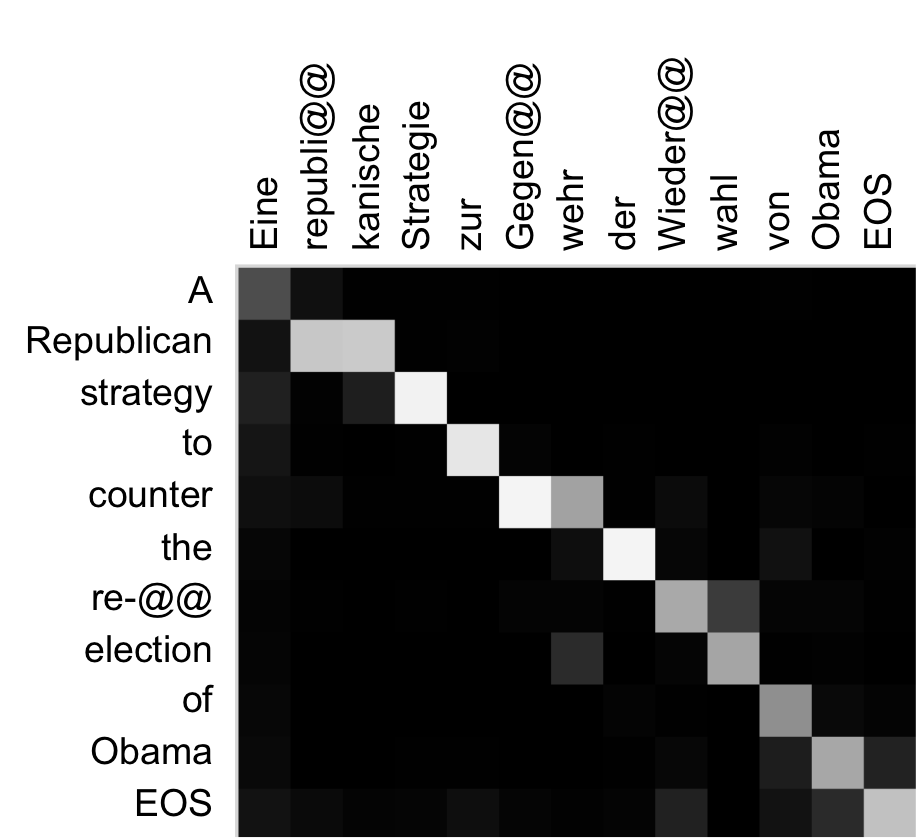}
        \includegraphics[height=1.9in]{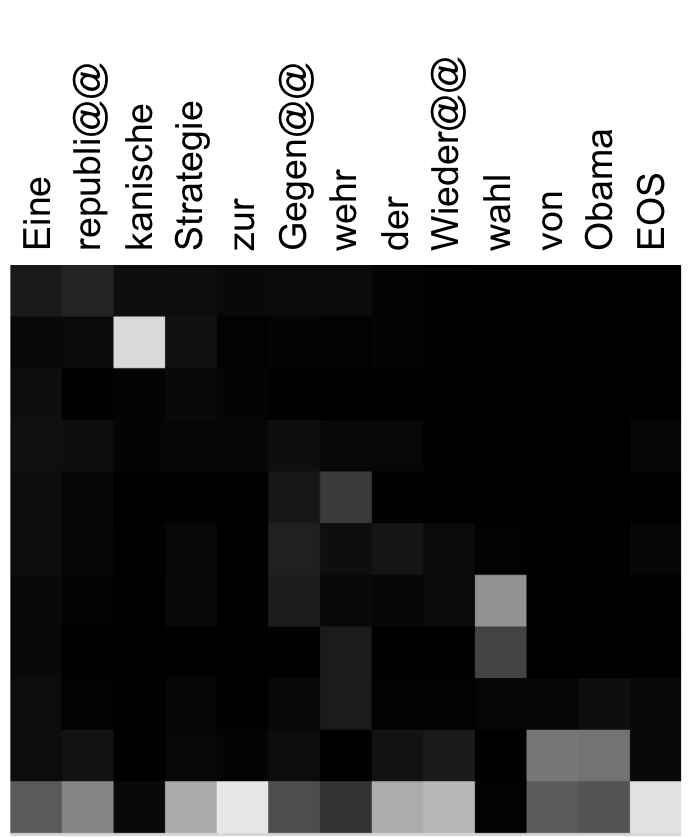}
        \includegraphics[height=1.9in]{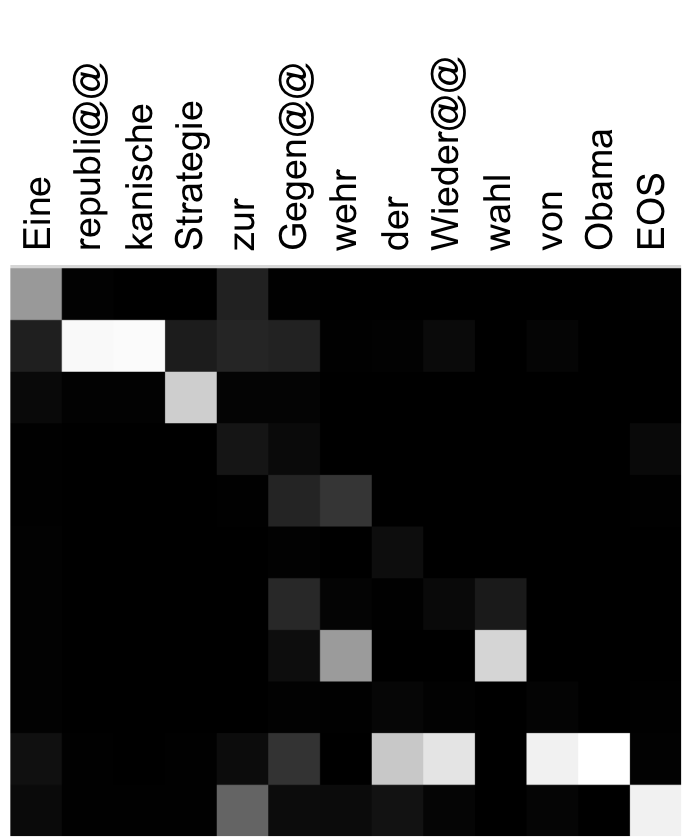}}

        \centering{\hspace{0.7in}(d) \hspace{1.40in} (e) \hspace{1.20in} (f)}

    \caption{Attention assignments in examplary dimensions with the fine-grained 
    attention model: attentions are focused on  
    (a) article (`a' and `the'), (b) preposition (`to' and `of'), (c) noun (`strategy', `election' and
    `Obama'), (d) the alignments, (e) adjective (`Republican' and `re-@@'), and (f) semantics
    words representing politics (`Republican', `strategy', `election' and 'Obama').}
\label{fig:att_ind}
\end{figure}

We can further inspect the alignment tensor $\alpha^d_{t',t}$ by visualizing the $d'$-th slice
of the tensor. 
Fig.~\ref{fig:att_ind} shows 6
example dimensions, where different dimensions focus on different perspective of 
translation. 
Some dimensions represent syntactic information, while others do semantic one. 
Also, syntactic information is handled in
different dimensions, according to the word type, like article (`a' and `the'),
preposition (`to' and `of'), noun (`strategy', `election' and `Obama'), and
adjective (`Republican' and `re-@@').  As semantic information,
Fig.~\ref{fig:att_ind}(f) shows a strong pattern of attention on the words
`Republican', 'strategy', `election' and `Obama', which seem to mean `politics'. 
Although we present one example of attention matrix, 
we observed the same patterns with other examples. 

\section{Conclusions}

In this paper, we proposed a fine-grained (or 2D) attention mechanism for 
neural machine translation. 
The experiments on En-De and En-Fi show that the proposed attention method improves the translation quality significantly. When the method was applied with the previous technique, contextualization, which was based on the similar idea, the performance was further improved. 
With alignment analysis, the fine-grained attention method revealed that 
the different dimensions of context play different roles in neural machine translation. 

We find it an interesting future work to
test the fine-grained attention with other NMT models like character-level models or multi-layered encode/decode models \cite{ling2015character,chung2016char}. Also, the fine-grained attention mechanism can be applied to different tasks like speech recognition. 

\section*{Acknowledgments}
The authors would like to thank the developers of Theano~\cite{Bastien2012}. 
This research was supported by Basic Science Research Program through 
the National Research Foundation of Korea(NRF) funded by the Ministry of Education (2017R1D1A1B03033341). Also, we acknowledge the support of the 
following agencies for research funding and computing support: 
NSERC, Calcul Qu\'{e}bec, Compute Canada, the Canada Research Chairs, 
CIFAR and Samsung. KC thanks the support by Facebook and Google 
(Google Faculty Award 2016). 



\begin{thebibliography}{25}
\expandafter\ifx\csname natexlab\endcsname\relax\def\natexlab#1{#1}\fi

\bibitem[Bahdanau et~al.(2015)Bahdanau, Cho, and Bengio]{Bahdanau2015}
Bahdanau, D., K.~Cho, and Y.~Bengio, 2015: {Neural Machine Translation by
  Jointly Learning to Align and Translate}. In {\em Proc. Int'l Conf. on
  Learning Representations (ICLR)\/}.

\bibitem[Bastien et~al.(2012)Bastien, Lamblin, Pascanu, Bergstra, Goodfellow,
  Bergeron, Bouchard, Warde-farley, and Bengio]{Bastien2012}
Bastien, F., P.~Lamblin, R.~Pascanu, J.~Bergstra, I.~Goodfellow, A.~Bergeron,
  N.~Bouchard, D.~Warde-farley, and Y.~Bengio, 2012: {Theano : new features and
  speed improvements}. In {\em NIPS 2012 deep learning workshop\/}.

\bibitem[Cho et~al.(2015)Cho, Courville, and Bengio]{cho2015describing}
Cho, K., A.~Courville, and Y.~Bengio, 2015: Describing multimedia content using
  attention-based encoder--decoder networks. {\em IEEE Transactions on
  Multimedia\/}, {\bf 17(11)}, 1875--1886.

\bibitem[Cho et~al.(2014)Cho, van Merrienboer, Bahdanau, and Bengio]{Cho2014}
Cho, K., B.~van Merrienboer, D.~Bahdanau, and Y.~Bengio, 2014: On the
  properties of neural machine translation: Encoder-decoder approaches. In {\em
  SSST-8, Eighth Workshop on Syntax, Semantics and Structure in Statistical
  Translation\/}, pp. 103--111.

\bibitem[Choi et~al.(2017)Choi, Cho, and Bengio]{hchoi2017csl}
Choi, H., K.~Cho, and Y.~Bengio, 2017: Context-dependent word representation
  for neural machine translation. {\em Computer Speech and Language\/}, {\bf
  45}, 149--160.

\bibitem[Chung et~al.(2016{\natexlab{a}})Chung, Cho, and Bengio]{chung2016char}
Chung, J., K.~Cho, and Y.~Bengio, 2016{\natexlab{a}}: A character-level decoder
  without explicit segmentation for neural machine translation. In {\em 54th
  Annual Meeting of the Association for Computational Linguistics\/}, pp.
  1693--1703.

\bibitem[Chung et~al.(2016{\natexlab{b}})Chung, Cho, and Bengio]{chung2016wmt}
---, 2016{\natexlab{b}}: Nyu-mila neural machine translation systems for
  wmt'16. In {\em The First Conference on Statistical Machine Translation
  (WMT)\/}.

\bibitem[Cohn et~al.(2016)Cohn, Hoang, Vymolova, Yao, Dyer, and
  Haffari]{cohn2016incorporating}
Cohn, T., C.~D.~V. Hoang, E.~Vymolova, K.~Yao, C.~Dyer, and G.~Haffari, 2016:
  Incorporating structural alignment biases into an attentional neural
  translation model. In {\em NAACL-HLT\/}, pp. 876--885.

\bibitem[Gulcehre et~al.(2015)Gulcehre, Firat, Xu, Cho, Barrault, Lin,
  Bougares, Schwenk, and Bengio]{gulcehre2015using}
Gulcehre, C., O.~Firat, K.~Xu, K.~Cho, L.~Barrault, H.-C. Lin, F.~Bougares,
  H.~Schwenk, and Y.~Bengio, 2015: On using monolingual corpora in neural
  machine translation. {\em arXiv preprint arXiv:1503.03535\/}.

\bibitem[Hochreiter and Schmidhuber(1997)]{Hochreiter1997}
Hochreiter, S. and J.~Schmidhuber, 1997: {Long short-term memory}. {\em Neural
  computation\/}, {\bf 9(8)}, 1735--80.

\bibitem[Jean et~al.(2015{\natexlab{a}})Jean, Cho, Memisevic, and
  Bengio]{Jean2015}
Jean, S., K.~Cho, R.~Memisevic, and Y.~Bengio, 2015{\natexlab{a}}: {On Using
  Very Large Target Vocabulary for Neural Machine Translation}. In {\em {53rd
  Annual Meeting of the Association for Computational Linguistics}\/}.

\bibitem[Jean et~al.(2015{\natexlab{b}})Jean, Firat, Cho, Memisevic, and
  Bengio]{jean2015montreal}
Jean, S., O.~Firat, K.~Cho, R.~Memisevic, and Y.~Bengio, 2015{\natexlab{b}}:
  Montreal neural machine translation systems for wmt15. In {\em Proceedings of
  the Tenth Workshop on Statistical Machine Translation\/}, pp. 134--140.

\bibitem[Kalchbrenner and Blunsom(2013)]{kalchbrenner2013recurrent}
Kalchbrenner, N. and P.~Blunsom, 2013: Recurrent continuous translation models.
  {\em EMNLP\/}, {\bf 3(39)}, 413.

\bibitem[Kingma and Ba(2014)]{Kingma2014adam}
Kingma, D.~P. and J.~L. Ba, 2014: Adam: A method for stochastic optimization.
  {\em arXiv preprint arXiv:1412.6980\/}.

\bibitem[Ling et~al.(2015)Ling, Trancoso, Dyer, and Black]{ling2015character}
Ling, W., I.~Trancoso, C.~Dyer, and A.~W. Black, 2015: Character-based neural
  machine translation. {\em arXiv preprint arXiv:1511.04586\/}.

\bibitem[Luong and Manning(2016)]{luong2016achieving}
Luong, M.-T. and C.~D. Manning, 2016: Achieving open vocabulary neural machine
  translation with hybrid word-character models. In {\em 54th Annual Meeting of
  the Association for Computational Linguistics\/}, p. 1054?1063.

\bibitem[Luong et~al.(2016)Luong, Pham, and Manning]{luong2016effective}
Luong, M.-T., H.~Pham, and C.~D. Manning, 2016: Effective approaches to
  attention-based neural machine translation. In {\em 2015 Conference on
  Empirical Methods in Natural Language Processing\/}, pp. 1412--1421.

\bibitem[Sennrich et~al.(2015{\natexlab{a}})Sennrich, Haddow, and
  Birch]{sennrich2015improving}
Sennrich, R., B.~Haddow, and A.~Birch, 2015{\natexlab{a}}: Improving neural
  machine translation models with monolingual data. {\em arXiv preprint
  arXiv:1511.06709\/}.

\bibitem[Sennrich et~al.(2015{\natexlab{b}})Sennrich, Haddow, and
  Birch]{Sennrich2015}
---, 2015{\natexlab{b}}: Neural machine translation of rare words with subword
  units. {\em arXiv preprint arXiv:1508.07909\/}.

\bibitem[Sennrich et~al.(2016)Sennrich, Haddow, and
  Birch]{sennrich2016edinburgh}
---, 2016: Edinburgh neural machine translation systems for wmt 16. In {\em The
  First Conference on Statistical Machine Translation (WMT)\/}.

\bibitem[Shen et~al.(2016)Shen, Cheng, He, He, Wu, Sun, and Liu]{Shen2016}
Shen, S., Y.~Cheng, Z.~He, W.~He, H.~Wu, M.~Sun, and Y.~Liu, 2016: {Minimum
  Risk Training for Neural Machine Translation}. In {\em 54th Annual Meeting of
  the Association for Computational Linguistics\/}, pp. 1683--1692.

\bibitem[Sutskever et~al.(2014)Sutskever, Vinyals, and Le]{Sutskever2014}
Sutskever, I., O.~Vinyals, and Q.~V. Le, 2014: {Sequence to Sequence Learning
  with Neural Networks}. In {\em Advances in Neural Information Processing
  Systems (NIPS)\/}.

\bibitem[Tu et~al.(2016)Tu, Lu, Liu, Liu, and Li]{tu2016modeling}
Tu, Z., Z.~Lu, Y.~Liu, X.~Liu, and H.~Li, 2016: Modeling coverage for neural
  machine translation. In {\em 54th Annual Meeting of the Association for
  Computational Linguistics\/}, pp. 76--85.

\bibitem[Van~der Maaten and Hinton(2012)]{van2012visualizing}
Van~der Maaten, L. and G.~Hinton, 2012: Visualizing non-metric similarities in
  multiple maps. {\em Machine learning\/}, {\bf 87(1)}, 33--55.

\bibitem[Xu et~al.(2015)Xu, Courville, Zemel, and Bengio]{Xu2015show}
Xu, K., A.~Courville, R.~Zemel, and Y.~Bengio, 2015: {Show, Attend and Tell :
  Neural Image Caption Generation with Visual Attention}. In {\em 32nd
  International Conference on Machine Learning\/}, pp. 2048--2057.

\end{thebibliography}

\end{document}